\documentclass[a4paper,twoside]{article}

\usepackage{epsfig}
\usepackage{subcaption}
\usepackage{calc}
\usepackage{amssymb}
\usepackage{amstext}
\usepackage{amsmath}
\usepackage{amsthm}
\usepackage{multicol}
\usepackage{pslatex}
\usepackage{apalike}
\usepackage{algorithm}
\usepackage{algpseudocode}
\usepackage{capt-of}
\usepackage{afterpage}
\usepackage{array}
\usepackage{amsmath}
\usepackage{amssymb}
\usepackage{graphicx}
\usepackage{booktabs} 
\usepackage{etoolbox}

\usepackage{SCITEPRESS}     

\begin{document}

\title{Multi-Label Network Classification via Weighted Personalized Factorizations}

\author{\authorname{Ahmed Rashed\sup{1}, Josif Grabocka\sup{1} and Lars Schmidt-Thieme\sup{1}}
\affiliation{\sup{1}Information Systems and Machine Learning Lab, University of Hildesheim, Hildesheim, Germany }
\email{\{ahmedrashed, josif, schmidt-thieme\}@ismll.uni-hildesheim.de}
}

\keywords{ Multi-Relational Learning, Network Representations, Multi-Label Classification, Recommender Systems, Document Classification}

\abstract{Multi-label network classification is a well-known task that is being used in a wide variety of web-based and non-web-based domains. It can be formalized as a multi-relational learning task for predicting nodes labels based on their relations within the network.
In sparse networks, this prediction task can be very challenging when only implicit feedback information is available such as in predicting user interests in social networks. Current approaches rely on learning per-node latent representations by utilizing the network structure, however, implicit feedback relations are naturally sparse and contain only positive observed feedbacks which mean that these approaches will treat all observed relations as equally important. This is not necessarily the case in real-world scenarios as implicit relations might have semantic weights which reflect the strength of those relations. If those weights can be approximated, the models can be trained to differentiate between strong and weak relations.
In this paper, we propose a weighted personalized two-stage multi-relational matrix factorization model with Bayesian personalized ranking loss for network classification that utilizes basic transitive node similarity function for weighting implicit feedback relations. Experiments show that the proposed model significantly outperforms the state-of-art models on three different real-world web-based datasets and a biology-based dataset.}

\onecolumn \maketitle \normalsize \vfill

\section{\uppercase{Introduction}}
\label{sec:introduction}

\noindent The classification of multi-label networks is one of the widely used tasks in network analysis such as in predicting or ranking user interests in social networks \cite{grover2016node2vec,perozzi2014deepwalk,krohn2012multi}, classifying documents \cite{tang2015line} in citation networks, predicting web-page categories in large network of websites and also in some biological domains such as predicting protein labels in protein-protein interaction networks \cite{grover2016node2vec}. 

The typical approach to predict node labels is by extracting a set of informative features from each node and train a classification model on them. This typical way of feature processing has two main significant drawbacks. First, to extract such informative features, one needs a prior expert domain knowledge to develop the features engineering process required to preprocess the raw data. Second, to extract useful features, a decent amount of raw information should be embedded with each node such as user profile details in social networks or document contents in citation networks. This kind of node embedded information might not always be available or accessible due to privacy settings.

Alternatively, current approaches rely on learning latent features for each node by analyzing the network structure and optimizing an objective function that will increase the accuracy of predicting node labels \cite{cai2018comprehensive}. Multi-relation matrix factorization \cite{krohn2012multi,jamali2010matrix,singh2008relational} is one famous example that follows this approach. This model represents network relations as matrices, and it factorizes the target relation matrix into two smaller matrices that represent the latent features of the interacting nodes. The main advantage of these alternative approaches is that they can be better generalized to almost all network classification tasks without any need for feature engineering or expert domain knowledge. However, they face significant challenges with very sparse networks and especially if these networks have implicit feedback relations. These implicit relations are dominantly very sparse, and relation edges are either observed or unobserved without explicit weights. In various real-world scenarios, those implicit relations have hidden semantic weights \cite{lerche2014using} which are not directly quantifiable but can be approximated using different weighting functions such as similarity measures. Current approaches that utilize the network structure to learn latent node representations fail to realize these hidden weights of implicit relations. A famous example of such relations is the friendship relation in social networks. This relation is a type of implicit feedback relations that models the interaction between nodes that have same types, and it is expressed as sparsely observed edges connecting those nodes. This kind of friendship relations frequently occurs in multi-relational settings and it is not only for representing a relation between users, but it can also represent a relation between any same type nodes. In networks data, all of the friendship observed edges would have the same importance weight while in real-life, some friendship relations are stronger than others. The real-life weights of such relations can be approximated by measuring the similarity between each two interacting nodes. The main advantage of using similarity is that it can be calculated using simple information from network structure such as nodes degrees without any need of complex auxiliary information which might not be available such as frequency interaction or timestamps. 

In this work, we introduce a similarity based personalized two-stage multi-relation matrix factorization model(Two-Stage-MR-BPR) for multi-label network classification and ranking. It utilizes the basic transitive node similarity for weighting implicit friendship relations and a two-stage training protocol to optimize the Bayesian personalized ranking loss. By optimizing the BPR loss, the model will output a ranked list of labels instead of only one label for any target node which means it will be suitable for recommender system problems and node classification problems if we just assume the top label as the predicted class.

The weighted Two-Stage-MR-BPR overcomes the drawbacks of MR-BPR and outperforms it in all of our experiments as it can distinguish between observed and unobserved relations along with learning the different strength weights of the observed relations.

Our contributions can be summarized as follows :
\begin{description}
 \item[$\bullet$]We utilize the transitive node similarity to approximate the semantic weights of all implicit relations that have interacting nodes of the same type to allow the Two-Stage-MR-BPR model to learn the strength weights of relations. 
 \item[$\bullet$]We propose a generalized two-stage learning algorithm that utilizes all available implicit and weighted relations for training the MR-BPR model. 
 \item[$\bullet$]We conduct multiple experiments on four real-world datasets. The results show that the proposed weighted Two-Stage-MR-BPR outperform the MR-BPR and current state-of-art models in multi-label and single-label classification problems. 
\end{description}

The rest of the paper is organized as follows. In Section 2, we summarize the related work. We discuss the problem formulation of the multi-label classification task in section 3. In section 4, we present and discuss the technical details of the Two-Stage-MR-BPR model. We present the experiential results in section 5. Finally, we conclude with discussing possible future work in section 6.

\section{\uppercase{Related Work}}

\noindent Current approaches for multi-label node classification automate the process of features extraction and engineering by directly learning latent features for each node.
These latent features are mainly generated based on the global network structure and the connectivity layout of each node. In earlier approaches such as \cite{tang2009relational,tang2009scalable}, they produce k latent features for each node by utilizing either the first k eigenvectors of a generated modularity matrix for the friendship relation \cite{tang2009relational} or a sparse k-means clustering of friendship edges \cite{tang2009scalable}. These k features are fed into an SVM for labels predictions.

Recently, semi-supervised \cite{thomas2016semi} and unsupervised approaches \cite{grover2016node2vec,perozzi2014deepwalk,yang2015network} have been proposed to extract latent node representations in networks data. These models are inspired by the novel approaches for learning latent representations of words such as the convolutional neural networks and the Skip-gram models \cite{mikolov2013efficient} in the domain of natural language processing. They formulate the network classification problem as discrete words classification problem by representing the network as a document and all nodes as a sequence of words. The Skip-gram can then be used to predict the most likely labels for each node based on the assumption that similar nodes will have same labels.

In \cite{krohn2012multi}, MR-BPR was proposed as learning to rank approach for tackling the multi-label classification problem by extending the BPR \cite{rendle2009bpr} model for multi-relational settings. This approach expresses the problem as a multi-relational matrix factorization trained to optimize the AUC measure using BPR loss. Each network relation is represented by a sparse matrix and the relation between nodes and labels will be the target being predicted. Because of the BPR loss, this model is considered suitable for sparse networks with implicit feedback relations; however, since all implicit feedback connections are only observed or unobserved, the MR-BPR fail to realize that some implicit links are stronger than others in real-life. To solve this drawback in the original single relation BPR model, \cite{lerche2014using} proposed BPR++ an extended version of the BPR model for user-item rating prediction. They utilized multiple weighting functions to approximate the logical weights between users and items. Those functions relied on the frequency of interaction and timestamps to weight each edge. In the training phase, they randomly alternate between learning to distinguish observed and unobserved relations and learning to rank weighted observed relations. This learning approach expands the BPR capacity to differentiate between strong and weak connections. 

Finally, our proposed Two-Stage-MR-BPR model follow a similar intuition to that of BPR++ and it considers the more general multi-relational settings which allow it to be used for any multi-label and single-label network classification problems. The proposed model utilizes the basic transitive node similarity for approximating the weights of implicit relations that have interacting nodes of the same type. In this work, we also propose MR-BPR++ an extended version of BPR++ for multi-relational settings. The learning algorithm for Two-stage-MR-BPR is different from MR-BPR++; it relies on two consecutive non-overlapping learning stages instead of random alternation. In the first stage, it allows the model to sufficiently learns to differentiate between strong and weak relations and in the second stage, it allows it to learns to differentiate between observed and unobserved relations.

\begin{figure*}
\includegraphics[width=\textwidth,height=3.8cm]{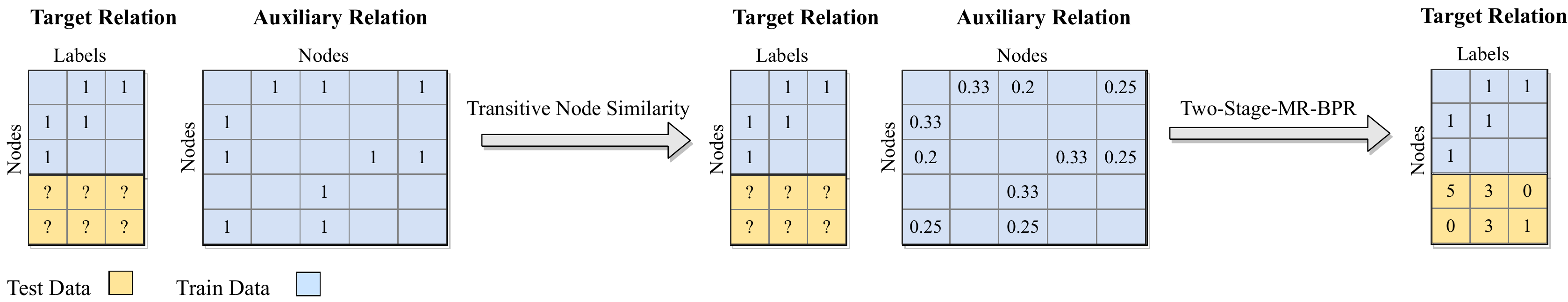}
\caption{Two-Stage-MR-BPR workflow. Initially, the transitive node similarity is used to weight all implicit relations with same type interacting nodes followed by learning a two stage multi-relational matrix factorization on all relations using the BPR loss to rank all labels with respect to each test node.}
\label{workflow}
\end{figure*}

\section{\uppercase{Problem Definition}}

\noindent The problem can be formulated similarly to \cite{krohn2012multi} as a relational learning setting on network data. Let G = (V,E) be a network where V is a set of heterogeneous nodes, and E is the set of edges. Each node can be seen as an entity and each edge represents a relation between two entities. Let $ \mathcal{N} :=\{N_1,N_2,...,N_{|\mathcal{N} |}\} $ be a set of node types and each type has a set of nodes as instances $N_i:=\{n_i^{(1)},n_i^{(2)},...,n_i^{(|N_i|)}\}$. Let $ \mathcal{R}:=\{R_1,R_2,...,R_{|\mathcal{R}|}\}$ be a set of relations and each relation represents interactions between two specific node types $N_{1R}$ and $N_{2R}$ such that $R\subseteq N_{1R} \times N_{2R}$.

Our primary task in this paper is to predict missing edges in a primary target relation $Y$, and all other relations will be considered auxiliary relations that can be used to improve the prediction accuracy. In multi-label network classification, the relation $Y$ represents the relation between a set of nodes and labels $Y\subseteq N_{Target} \times N_{Label}$, such as the relationship between multiple interests and users in social networks or document-labels and documents in citation networks. Examples of auxiliary relations are the friendship relation in social networks or citation links in citation networks. 

The task of predicting missing edges in the target relation can be formulated as a ranking problem where we try to drive a ranked list of labels that represent the likelihood that a specific node belongs to each of them. 

In case of sparse auxiliary relations with implicit edges, the current multi-relational matrix factorization model with BPR loss \cite{krohn2012multi} does not exploit the full potential of the BPR loss because it only distinguishes between observed and unobserved edges without considering the edges weights \cite{lerche2014using}.

Our proposed approach addresses the shortcomings of the current multi-relational matrix factorization model by firstly using transitive node similarity for weighting the implicit auxiliary relations and using a Two-Stage-MR-BPR learning algorithm that can rank observed edges and distinguish between observed and unobserved edges. The proposed approach is also suitable for cold-start scenarios where only the auxiliary relations information are available for the target nodes.
\section{\uppercase{Proposed model}}
\noindent The proposed model can be formulated as a two-stage multi-relational matrix factorization using basic transitive node similarity for weighting implicit relations. Initially, the transitive node similarity is used to weight all observed edges in any implicit auxiliary relations that has interacting nodes of the same type. In the first stage of learning, the model is trained to rank edges based on their weights. In the second stage, the model is trained to differentiate between observed and unobserved edges. Figure \ref{workflow} illustrates the workflow of the Two-Stage-MR-BPR model and each step will be discussed in details in the following subsections.

\subsection{Basic Transitive Node Similarity for Implicit Feedback Relations}
In the relational learning setting, in order to apply the two-stage MR-BPR learning technique we need first to convert all possible implicit feedback relations into weighted relations. To convert implicit relations into weighted relations, one needs a suitable weighting function that approximate relation weights by utilizing the available embedded information in each relation such as frequency or timestamps of interactions in user-item relations, or similarity measures in friendship relations. In the most basic case where there is no available embedded information, an implicit relation can be considered a relation that was weighted by a constant weighting function which outputs only one value if it encounters an observed edge.

Friendship relations are one of the prominent types of implicit relations in networks data. It can represent any relation between nodes of the same type such as users friendships in social networks or web links between web pages or citation links between documents. For weighting edges in general friendship relations, the similarity measures such as Adamic/Adar, common neighbors, Jaccard index and friend transitive node similarity FriendTNS \cite{symeonidis2010transitive} are considered the best candidates to act as weighting functions \cite{symeonidis2010transitive,liben2007link}. In our proposed approach we used the FriendTNS function because it provided superior accuracy over other similarity functions \cite{symeonidis2010transitive,ahmed2016supervised} in link prediction tasks and it requires minimum auxiliary information from the network structure. FriendTNS was used for weighting all the observed edges in all available implicit friendship relations and it was calculated only for observed edges because it is computationally expensive to calculate weights for all possible node pairs in very sparse networks. The FriendTNS similarity between two nodes is calculated using equation (1).
\begin{equation}
 FTNS(n_i,n_j):= 
\begin{cases}
    \frac{1}{deg(n_i)+deg(n_j)-1},&\text{if }(n_i,n_j)\in R\\
    0,              &\text{if }(n_i,n_j) \notin R\\
\end{cases}
\end{equation}

\noindent where $deg(n_i)$ and $deg(n_j)$ is the degree of nodes $n_i$ and $n_j$ respectively. In case of directed graphs, we used the summation of the node's in-degree and out-degree as the total degree. 

\subsection{ Multi-Relational Matrix Factorization With Basic Transitive Node Similarity }
To formulate the problem as a multi-relational matrix factorization, each node type $N_i$ can be represented by a matrix $E_i \in \mathbb{R}^{|N_i| \times k}$ where the rows are the latent feature vectors for all instances in the node type, and $k$ represents the number of latent factors defined in the model. Similarly, each implicit relation $R$ can be represented by a matrix $R \in \mathbb{R}^{|N_{1R}| \times |N_{2R}|}$ where $N_{1R}$ and $N_{2R}$ are the two types of the interacting nodes inside relation $R$. Each entry in the relation matrix is given by

\begin{equation}
\begin{aligned}
R(n_{1R}^{(i)},n_{2R}^{(j)}) :=
\begin{cases}
    Weight(n_{1R}^{(i)},n_{2R}^{(j)}),&\text{if}(n_{1R}^{(i)},n_{2R}^{(j)}) \in R  \\ \\
    unobserved,              &\text{if}(n_{1R}^{(i)},n_{2R}^{(j)}) \notin R\\ 
\end{cases}
\end{aligned}
\end{equation}
\noindent where $Weight(n_{1R}^{(i)},n_{2R}^{(j)})$ is the weighting function used to approximate the weights of implicit relation between any two nodes such as similarity functions in case of friendship relations or frequency of interaction in case of user-item relations. If no available embedded information can be used to weight relations, a constant weighting function is assumed.

Finally, each relation $R$ can be approximated by multiplying the latent matrices of the two relation node types $E_{1R}$ and $E_{2R}$ such that $R \approx E_{1R} \times E_{2R}^{T} $. For simplicity, we define a set of all the model parameters $\Theta :=\{E_1,E_2,...,E_{|\mathcal{N} |}\}$ which contain the matrices of all nodes types and our general objective will be to find the set of matrices $\Theta$ that minimize the sum of losses over all relations. 

\subsection{Two-Stage MR-BPR}
The original BPR model \cite{rendle2009bpr} assumes that for a given user $u$, any item $i$ this user interacted with should be ranked higher than any item $j$ he did not interact with. In order to do so, the BPR model learns to maximize the difference $\hat{x}_{u,i,j}^{R}$ between the predicted rating $\hat{r}(u,i)$ for an observed item $i$ and the rating $\hat{r}(u,j)$ for an unobserved item $j$. 

To follow the same notation in a multi-relational setting, for any given relation $R$, the user $u$ will represent a node of type $N_{1R}$, while $i$ and $j$ will represents two nodes of type $N_{2R}$. For each relation $R$ the baseline MR-BPR model samples a set of triples $D_R$ which is defined as follows:
\begin{displaymath}
\begin{aligned}
D_R := \{(u,i,j)| (u,i) \in R \wedge (u,j) \notin R \}
\end{aligned}
\end{displaymath}

The sampling is done using bootstrap sampling with replacement. The model is then trained to maximize the difference between the predicted ratings of the observed edges and unobserved edges for all relation using equations (2) and (3).
 
\begin{equation}
\textrm{BPR\mbox{-}Opt}(R,E_{1R} {E_{2R}}^T)= \sum_{(u,i,j)\in R} ln\sigma(\hat{x}_{u,i,j}^{R})
\end{equation}
\begin{equation}
\begin{aligned}
\textrm{MR\mbox{-}BPR}(R,\Theta)=& \sum_{R \in \mathcal{R}} \alpha_R \textrm{BPR\mbox{-}Opt}(R,E_{1R} {E_{2R}}^T) \\ &+ \sum_{E \in \Theta} \lambda_E ||E||^2
\end{aligned}
\end{equation}

\noindent where $\sigma$ is the sigmoid logistic function and $\alpha_{R}$ is the loss weight for relation $R$. 
By following this learning approach, the MR-BPR model learns to distinguish between observed edges and unobserved edges over iterations. This approach is not optimal as it fails to realize the different semantic weights of the implicit relations. In \cite{lerche2014using} they proposed a new learning technique for the original BPR model called BPR++ which extend the BPR to learn weighted relations. There proposed extension allow the BPR model to provide better rankings for item ratings by utilizing the timestamps and frequency of user interactions to weight user-item edges. BPR++ is randomly alternating between learning to rank observed weighted edges and learning to distinguish between observed and unobserved edges. Instead of using such random alternation between the two learning tasks for multi-relational settings, we propose a two-stage learning approach that decouple the two learning tasks and learn them sequentially to avoid information overwrites across iterations. 
When applied to multi-relational settings, the proposed two-stage learning protocol and BPR++ will utilize a separate set of triples $D_{R}^{++}$ beside the original set $D_{R}$. This new set contains observed weighted edges sampled using bootstrap sampling with replacement for each available weighted relation as follows:

\begin{displaymath}
\begin{aligned}
D_{R}^{++} := \{(u,i,j)|Weight(u,i)>Weight(u,j)\wedge \\(u,i)\in R \wedge(u,j) \in R \} 
\end{aligned}
\end{displaymath}

The main difference between the proposed two-stage learning protocol and BPR++ is that the later will rely on random alternating sampling from $D_{R}$ and $D_{R}^{++}$ which introduce the risk of having information loss as some iterations might overwrite the previously learned information, e.g. if an node was selected first as an observed item from $D_{R}$ and in the next iteration it was selected as the lower weighted item from $D_{R}^{++}$, the second iteration will overwrite the information gained in the first iteration as it will decrease the score of the item after it has been increased. On the other hand, Two-Stage-MR-BPR overcome such problem by learning to rank all weighted edges first then it learns to distinguish observed and unobserved edges afterward with no overlap between the two stages in each epoch. This means that the second stage will shift the learned scores of the observed edges away from the unobserved ones while maintaining the learned rankings between the weighted observed edges.
\begin{figure}[!ht]
  \begin{algorithmic}[1]
    \Procedure {Two-Stage-MR-BPR}{$\mathcal{D},\mathcal{R},\Theta$}
      \State Initialize All $E \in \Theta$
      \Repeat
        \For{ $R \in \mathcal{R}$ }
			 \State // (Stage One)
 			\If{$D_{R}^{++}\backslash D_{R}\neq \phi$ }        		
               \For{ $ObsEdges_R$ times}          		
                 	\State  \strut draw (u,i,j) from $D_{R}^{++}$
                  	\State  \strut $\Theta \gets \Theta + \mu\frac{\partial( \textrm{MR\mbox{-}BPR}(R,\Theta))}{\partial\Theta}$ 
                 \EndFor
            \EndIf 
             \State // (Stage Two)
           		 \For{ $ObsEdges_R$ times}       
                 	\State draw (u,i,j) from $D_{R}$
                     \State $\Theta \gets \Theta + \mu\frac{\partial( \textrm{MR\mbox{-}BPR}(R,\Theta))}{\partial\Theta}$ 
                 \EndFor         
       \EndFor
      \Until{convergence}\\
      \Return{$\Theta$}
    \EndProcedure
  \end{algorithmic}
  \caption{Two-Stage\mbox{-}MR\mbox{-}BPR algorithm with learning rate $\mu$ and L2 regularization $\lambda_{\Theta}$ } 
  \label{tsmrbpr}
\end{figure}

\begin{figure}[!ht]
  \begin{algorithmic}[1]
    \Procedure{MR-BPR++}{$\mathcal{D},\mathcal{R},\Theta$}
      \State Initialize All $E \in \Theta$
      \Repeat
        \For{ $R \in \mathcal{R}$ }     	
        			\State $r= random (0,1)$
                    \If{$r \le \beta \wedge D_{R}^{++}\backslash D_{R}\neq \phi$ }
                    \State draw (u,i,j) from $D_{R}^{++}$
                    \Else
                 	\State draw (u,i,j) from $D_{R}$
                    \EndIf 
                     \State $\Theta \gets \Theta + \mu\frac{\partial( \textrm{MR\mbox{-}BPR}(R,\Theta))}{\partial\Theta}$                               
       \EndFor
      \Until{convergence}\\
      \Return{$\Theta$}
    \EndProcedure
  \end{algorithmic}
  \caption{MR\mbox{-}BPR++ algorithm with learning rate $\mu$, probability threshold $\beta$ and L2 regularization $\lambda_{\Theta}$ } 
  \label{mrbprplusplus}
\end{figure}

In our experiments, we applied both learning protocols on the MR-BPR model for performance comparison and we used the basic FriendTNS similarity as weighting function for all implicit relations where the participating nodes have the same type. In each learning epoch during the training phase, the number of sampling steps for $D_{R}^{++}$ and $D_{R}$ is equal to the number of observed edges $ObsEdges_R$ in $R$, similar to the original BPR \cite{rendle2009bpr} and MR-BPR \cite{krohn2012multi} models. The generalized algorithms for training the Two-Stage-MR-BPR and MR-BPR++ are described in Figures \ref{tsmrbpr} and \ref{mrbprplusplus}.

In the experiments section, we compared the two models against each other and the original MR-BPR model. The results showed that the proposed two-stage model provides better accuracy for multi-relational settings where we have multiple sparse relations without timestamps or frequency of interactions that can be used to weight relations. 
\begin{table*}[!ht]
  \caption{Datasets Statistics}
  \label{tab:datasets}
  \centering
  \begin{tabular}{ccccccc}
    \toprule
    & Type  & Nodes  & Labels  & Edges & Features & Sparsity \\
    \midrule
    BlogCatalog &Undirected  &10312 & 39 &  667966 &-& 99.37\%\\            
    PPI&Undirected  &3890 & 50 &  76584 &-& 99.49\%\\  
    Wiki&Directed  &2405 & 19 &  17981 &-& 99.68\%\\   
    Cora&Directed  &2708 & 7 &  5429 &1433& 99.92\%\\ 
    \bottomrule
  \end{tabular}
\end{table*}

\section{\uppercase{Experiments}}

\subsection{Datasets}
We applied the Two-Stage MR-BPR and MR-BPR++ on four network classification datasets from four different domains. The first three datasets contain two relations while the fourth dataset is a citation network where nodes have an embedded feature vector that can be considered as third relations.
\begin{description}
  \item[$\bullet$] BlogCatalog \cite{zafarani2009social} : This dataset represents a large social network from the BlogCatalog website. It has two relations, a target relation which represents the relation between groups and users, and an auxiliary relation representing the friendship between users.
  \item[$\bullet$] Wiki \cite{tu2016max} : This dataset represents a network of Wikipedia web pages. It also has two relations, a target relation which represents the categories of the web pages, and an auxiliary relation that represents links between the web pages.
    \item[$\bullet$] Protein-Protein Interactions (PPI) \cite{breitkreutz2007biogrid} : This dataset is a network of protein-protein interactions for homo sapiens. It has two relations, a target one which represents the relation between protein-labels and proteins, and an auxiliary relation that represents the interactions between proteins and other proteins. This dataset was used to check how well our proposed model performs in non-web-based domains. 
    \item[$\bullet$] Cora \cite{sen2008collective} : This dataset represents a citation network where each document has 1433 binary feature vector representing words occurrence. This dataset can be considered as having three relations, a target relation which represents the class label of a document, an auxiliary relation that represents citation links between documents and a final auxiliary relation that represents a relation between a document and words that exist in this document. 
\end{description}
Table \ref{tab:datasets} shows the detailed statistics of the datasets.

\subsection{Baselines}
\begin{description}
  \item[$\bullet$] MR-BPR \cite{krohn2012multi} : The original MR-BPR model that utilizes implicit auxiliary relations for ranking node labels. This model does not utilize transitive node similarities.
  \item[$\bullet$] DeepWalk \cite{perozzi2014deepwalk} : One of the well-known models for multi-label network classification. This model learns node latent representations by utilizing uniform random walks in the network.  
   \item[$\bullet$] Node2Vec \cite{grover2016node2vec} : This is one of the state-of-art models for multi-label network classification and can be seen as a generalized version of DeepWalk with two guiding parameters $p$ and $q$ for the random walks.
    \item[$\bullet$] GCN \cite{thomas2016semi} : This model is one of the state-of-art models for document classification in citation networks. It relies on multi-layered graph convolutional neural network for learning network representation with text features.
    \item[$\bullet$] TADW \cite{yang2015network} :This model is also one of the state-of-art models for document classification in citation networks. It is an extended version of the original Deep Walk model for learning network representation with text features.

\end{description}
On Cora dataset, our proposed model was compared only against GCN and TADW because they require nodes with embedded textual features which is missing in the first three datasets. On the other hand DeepWalk and Node2Vec where not used on Cora dataset because they can't represent nodes with embedded features.
 \newrobustcmd{\B}{\bfseries}


\newcolumntype{H}{>{\setbox0=\hbox\bgroup}c<{\egroup}@{}}
\begin{table*}[!ht]
  \caption{Mutli-lable classification results on BlogCatalog dataset}
  \label{tab:blog}
  \centering
  \begin{tabular}{llcHcHcHcHc}
    \toprule
    & \%Lable Nodes  & 10\%  & 20\%  & 30\% & 40\%  & 50\% & 60\% & 70\% & 80\% & 90\%\\
    \midrule
    Micro-F1(\%)         
 	  &DeepWalk  &33.71&36.67&38.22&39.20&39.37&40.01&40.64&41.04&41.11\\  
    &Node2Vec  &33.72&36.91&38.33&39.42&39.98&40.52&40.75&41.82&42.16 \\  
    &MR-BPR  &36.16&37.75&39.24&40.07&40.68&40.29&40.39&41.24&40.64\\      
    &MR-BPR++  &35.47&38.00&39.49&40.22&40.84&40.83&40.99&41.92&41.14\\  
    &Two-Stage-MR-BPR  &\B 37.27**&\B39.30**&\B40.49**&\B41.52*&\B42.22**&\B41.76**&\B42.03**&\B42.83**&\B42.51*\\   
    \midrule
    Macro-F1(\%)       
    &DeepWalk  &18.19&22.18&23.61&24.63&25.32&26.24&27.20&27.21&27.84\\  
    &Node2Vec  &19.24&23.13&24.70&25.64&26.80&27.81&27.80&28.75&29.15\\  
    &MR-BPR  &22.21&24.74&26.20&27.59&27.95&28.26&28.49&29.33&29.13\\      
    &MR-BPR++  &22.03&25.21&26.55&27.89&28.19&28.65&29.12&29.53&29.62\\  
    &Two-Stage-MR-BPR  &\B23.18**&\B25.66**&\B26.91**&\B28.16**&\B28.69**&\B28.76&\B29.54**&\B29.85&\B30.55*\\  
    \bottomrule
  \end{tabular}
  \\ \footnotesize Significantly outperforms MR-BPR at the: ** 0.01 and * 0.05 levels. 
  
\end{table*}

\begin{table*}[!ht]
  \caption{Mutli-lable classification results on PPI dataset}
  \label{tab:ppi}
  \centering
  \begin{tabular}{llcHcHcHcHc}
    \toprule
    & \%Lable Nodes  & 10\%  & 20\%  & 30\% & 40\%  & 50\% & 60\% & 70\% & 80\% & 90\%\\
    \midrule
    Micro-F1(\%)         
 	&DeepWalk  &15.89&17.77&18.68&19.39&20.85&21.75&22.35&22.70&24.11\\  
    &Node2Vec  &15.09&16.89&17.52&19.00&20.38&21.43&22.02&22.25&22.65\\  
    &MR-BPR  &17.11&19.68&20.87&21.87&22.61&22.73&23.55&23.08&23.44\\      
    &MR-BPR++  &16.97&19.46&20.62&21.90&22.46&23.14&23.33&23.28&23.56\\  
    &Two-Stage-MR-BPR  &\B18.21**&\B20.45**&\B21.88**&\B22.63**&\B23.31**&\B23.70**&\B24.78**&\B24.48**&\B25.38**\\   
    \midrule
    Macro-F1(\%)       
    &DeepWalk  &12.73&14.20&15.41&17.06&18.50&18.84&18.49&18.49&19.15\\  
    &Node2Vec  &12.17&13.47&14.51&16.72&18.01&18.62&18.89&18.45&18.76\\  
    &MR-BPR  &12.88&15.56&16.83&18.09&18.81&18.98&19.54&19.16&19.48\\      
    &MR-BPR++  &12.71&15.40&16.58&18.00&18.66&19.37&19.54&19.46&19.66\\  
    &Two-Stage-MR-BPR  &\B13.96**&\B16.31**&\B17.96**&\B18.75**&\B19.43**&\B19.99**&\B20.72**&\B20.51**&\B21.41**\\  
    \bottomrule
  \end{tabular}
  \\ \footnotesize Significantly outperforms MR-BPR at the: ** 0.01 and * 0.05 levels. 
  
\end{table*}
\subsection{Experimental Protocol and Evaluation}
We followed the same experimental protocol in \cite{krohn2012multi,perozzi2014deepwalk,grover2016node2vec}. We used 10-fold cross-validation experiments on each target relation. These experiments were applied using different percentages of labeled nodes ranging from 10\% to 90\%. In each experiment, we only used the defined percentage of labeled nodes for training along with all the auxiliary relations, while the remaining percent of nodes were used for testing. We used Micro-F1 and Macro-F1 measures for performance evaluation on Blog Catalog, PPI and Wiki Dataset, and Accuracy on Cora Dataset. 

We used the same hyper-parameters that were used in the original baselines' papers, and grid-search was used to find the best hyper-parameters if none were mentioned for the target dataset.

\subsection{Results} 
\begin{figure*}[!ht]
\includegraphics[width=\textwidth,height=13cm]{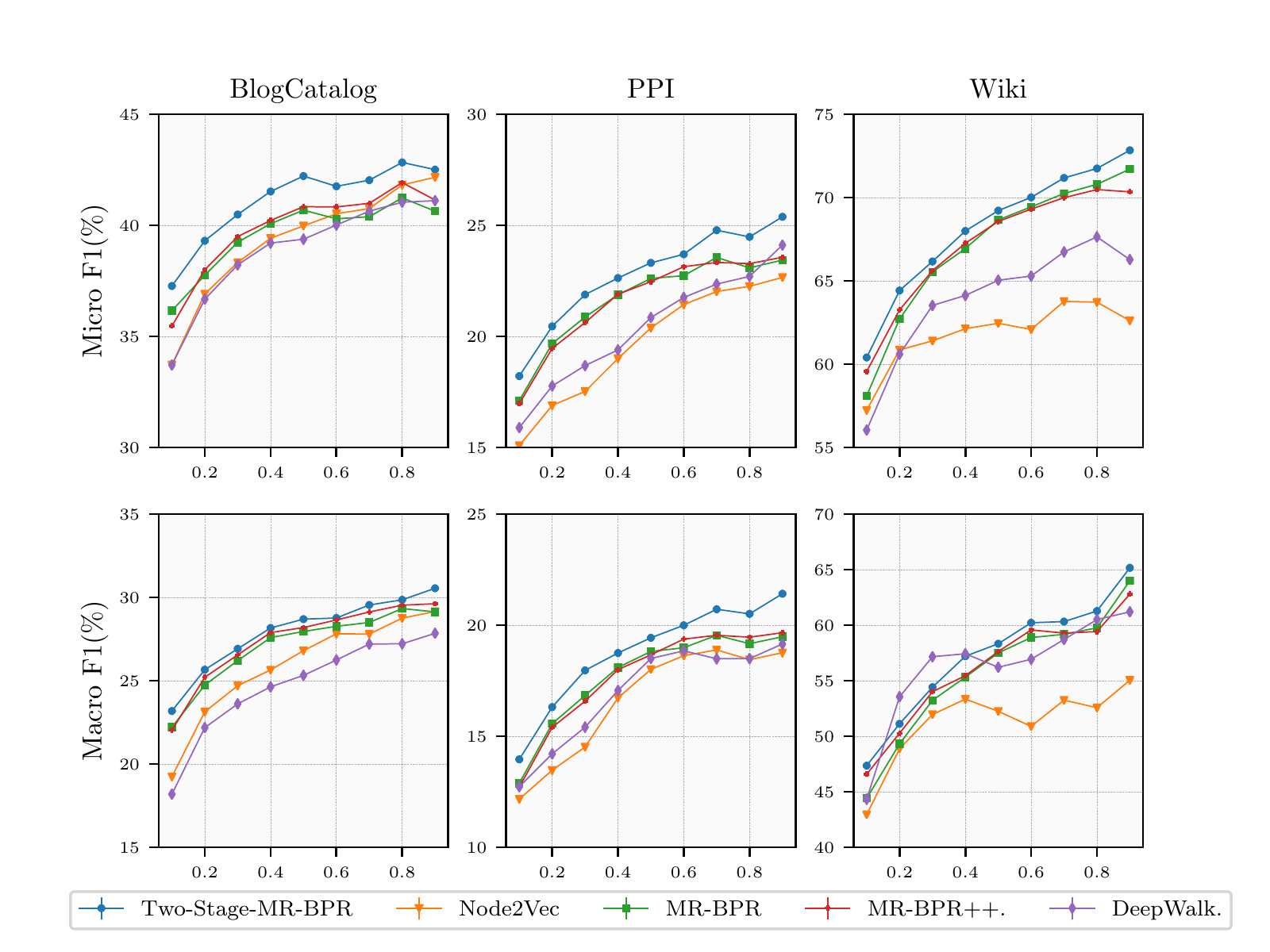}
\caption{Evaluation of the Two-Stage-MR-BPR model against other baseline models. The $x$ axis denotes the percent of labeled nodes used in the training phase, while the $y$ axis denotes the Micro-F1 and Macro-F1 scores.}
\label{results}
\end{figure*}

The experimental results on the four datasets are shown in Tables \ref{tab:blog}, \ref{tab:ppi}, \ref{tab:wiki}, \ref{tab:Cora}, and Figure \ref{results}. The results shows that Two-Stage-MR-BPR with transitive node similarity outperformed the original MR-BRP model in all train-splits.
In comparison with other well-known models for multi-label network classification, the Two-Stage-MR-BPR model outperformed the state-of-art Node2Vec over all trains-splits on BlogCatalog, PPI and Wiki datasets. It is worthy to note that all improvements over Node2Vec are statistically significant with a p-value less than 0.01 using paired t-test. Two-Stage-MR-BPR also outperformed DeepWalk over all trains-splits on BlogCatalog and PPI, while on Wiki, DeepWalk only achieved better Macro-F1 scores on the 20\%, 30\%, and 40\% trains-splits.
The results also show that Two-Stage-MR-BPR outperformed all other models in terms of Micro-F1 with 40\% less data using the 50\% train-split on the BlogCatalog dataset. On PPI, It outperformed all other models in terms of Micro-F1 and Macro-F1 with 30\% less data using the 60\% train-split.
On the other hand, the document classification results on the Cora datasets show that Two-Stage-MR-BPR outperformed the state-of-art models on 10\% and 90\% splits, while it achieves comparable results on the 50\% train-splits.

\begin{table*}[!ht]
  \caption{Mutli-lable classification results on Wiki dataset}
  \label{tab:wiki}
  \centering
  \begin{tabular}{llcHcHcHcHc}
    \toprule
    & \%Lable Nodes  & 10\%  & 20\%  & 30\% & 40\%  & 50\% & 60\% & 70\% & 80\% & 90\%\\
    \midrule
    Micro-F1(\%)         
 	&DeepWalk  &56.04&60.60&63.52&64.13&65.03&65.29&66.73&67.64&66.27\\  
    &Node2Vec  &57.24&60.85&61.40&62.13&62.45&62.08&63.76&63.72&62.61 \\  
    &MR-BPR  &58.10&62.71&65.54&66.94&68.66&69.44&70.24&70.79&71.71\\      
    &MR-BPR++  &59.56&63.26&65.61&67.26&68.56&69.31&69.99&70.48&70.34\\  
    &Two-Stage-MR-BPR  &\B60.40**&\B64.42**&\B66.16&\B67.99**&\B69.21&\B70.01&\B71.18&\B71.74&\B72.84\\   
    \midrule
    Macro-F1(\%)       
    &DeepWalk  &44.33&\B53.55&\B57.16&\B57.42&56.21&56.93&58.71&60.50&61.20\\  
    &Node2Vec  &42.95&48.88&51.96&53.34&52.25&50.88&53.25&52.57&55.04\\  
    &MR-BPR  &44.41&49.32&53.23&55.32&57.50&58.88&59.17&59.76&63.99\\      
    &MR-BPR++  &46.58&50.27&54.01&55.45&57.62&59.57&59.29&59.42&62.79\\  
    &Two-Stage-MR-BPR  &\B47.35**&51.12**&54.41*&57.20*&\B58.33&\B60.21&\B60.32&\B61.27&\B65.16\\  
    \bottomrule
  \end{tabular}
  \\ \footnotesize Significantly outperforms MR-BPR at the: ** 0.01 and * 0.05 levels. 
\end{table*}

In comparison with Two-Stage-MR-BPR, MR-BRP++ also outperformed MR-BRP on BlogCatalog over most of the train splits, but it had minimal performance gains in some train-splits on PPI, Wiki and Cora. These results demonstrate the importance of using transitive node similarity to weight implicit relations, and they show that using two sequential non-overlapping stages to train the BPR loss is better than randomly alternating between ranking observed edges and distinguishing them from unobserved edges.  

\begin{table*}[!ht]
  \caption{Document classification results on Cora dataset}
  \label{tab:Cora}
  \centering
  \begin{tabular}{llccc}
    \toprule
    & \%Lable Nodes  & 10\% & 50\%  & 90\%\\
    \midrule
    Accuracy(\%)         
 	&GCN  &78.37 & \B86.53 & 86.39\\  
    &TADW   & 75.24& 85.99& 85.60\\  
    &MR-BPR  & 75.03 &78.76 &81.66\\      
    &MR-BPR++   & 76.24&78.10 &81.10\\  
    &Two-Stage-MR-BPR  & \B79.30** & 84.20 &\B86.86\\   
    \bottomrule
  \end{tabular}
  \\  \footnotesize Significantly outperforms GCN at the: ** 0.01 and * 0.05 levels. 
\end{table*}

\subsection{Reproducibility of the Experiments} 
For each model we used the following hyper-parameters during in experiments.
\begin{description}
 \item[$\bullet$] MR-BPR: The hyper-parameters are $k = 500$, $\mu = 0.02$, $\lambda_{user} = 0.0125$, $\lambda_{item} = 0.0005$, 300 iterations and $\alpha = 0.5$ for BlogCatalog; $k = 500$, $\mu = 0.01$, $\lambda_{protein} = 0.0125$, $\lambda_{label} = 0.0005$, 400 iterations and $\alpha = 0.5$ for PPI; $k = 600$, $\mu = 0.02$, $\lambda_{page} = 0.0125$, $\lambda_{label} = 0.0005$, 1000 iterations and $\alpha = 0.5$ for Wiki; and $k = 900$, $\mu = 0.03$, $\lambda_{document} =0.005$, $\lambda_{label} = 0.0001$, $\lambda_{words} = 0.0001$, 1400 iterations and $\alpha = 0.33$ for Cora.
 
 \item[$\bullet$] MR-BPR++: The hyper-parameters are $\beta = 0.75$, $k = 500$, $\mu = 0.02$, $\lambda_{user} = 0.0125$, $\lambda_{item} = 0.0005$, 300 iterations and $\alpha = 0.5$ for BlogCatalog; $\beta = 0.75$, $k = 500$, $\mu = 0.01$, $\lambda_{protein} = 0.0125$, $\lambda_{label} = 0.0005$, 400 iterations and $\alpha = 0.5$ for PPI; $\beta = 0.75$, $k = 600$, $\mu = 0.02$, $\lambda_{page} = 0.0125$, $\lambda_{label} = 0.0005$, 1000 iterations and $\alpha = 0.5$ for Wiki; and $k = 900$, $\mu = 0.03$, $\lambda_{document} =0.005$, $\lambda_{label} = 0.0001$, $\lambda_{words} = 0.0001$, 1400 iterations and $\alpha = 0.33$ for Cora.
 \item[$\bullet$] Two-Stage-MR-BPR: We used the same hyper-parameters of MR-BPR
 \item[$\bullet$] DeepWalk: The hyper-parameters are $d = 128$, $r = 10$, $l = 80$ and $k = 10$ for all datasets.
 \item[$\bullet$] Node2Vec: The hyper-parameters are $d = 128$, $r = 10$, $l = 80$, $k = 10$, $p = 0.25$ and $q= 0.25$ for BlogCatalog; $d = 128$, $r = 10$, $l = 80$, $k = 10$, $p = 4$ and $q= 1$ for PPI and Wiki.
  \item[$\bullet$] GCN: We used the same hyper-parameters from the original paper. Dropout rate  $= 0.5$, L2 regularization = $5.10^{-4}$ and 16 (number of hidden units)
  \vfill
 \item[$\bullet$] TADW: We used the same default hyper-parameters from the original papers which are $k = 80$ and $\lambda = 0.2$ 
\end{description}

\section{\uppercase{Conclusions}}
\noindent In this paper, we proposed the similarity based Two-Stage-MR-BPR model that exploits the full potential of the Bayesian personalized ranking as a loss function. Two-Stage-MR-BPR relies on converting all implicit feedback relations into weighted relations using basic transitive node similarity to approximate the semantic weights of relations in the real world. This conversion step allows the Two-Stage-MR-BPR to learn ranking edges based on their weights and to learn to distinguish observed edges from unobserved ones. Experiments on four real-world datasets showed that Two-Stage-MR-BPR outperformed the original MR-BPR and other state-of-art models in the task of multi-label network classification and document classification. In future work, we are planning to explore and integrate several possible approaches to further expand the capacity of the BPR models. Two of those possible approaches for expansion are learning nonlinear latent representation and using non-uniform samplers \cite{rendle2014improving}.

\section*{\uppercase{Acknowledgements}}

\noindent The authors gratefully acknowledge the funding of their work through the {\"U}berDax project (https://www.ismll.uni-hildesheim.de/projekte/ UberDax.html) which is sponsored by the Bundesministerium für Bildung und Forschung (BMBF) under the grant agreement no. 01IS17053.

\bibliographystyle{apalike}
{\small
\bibliography{example}}

\end{document}